%% file: main.tex
\pdfoutput=1

\documentclass[11pt]{article}

\usepackage[final]{acl}

\usepackage{times}
\usepackage{latexsym}
\usepackage{booktabs}
\input{math_commands}

\usepackage{amsmath,amsfonts,amssymb}
\usepackage{bbm}

\usepackage{amsthm}
\theoremstyle{definition}

\newtheorem{quest}{Question}
\theoremstyle{remark}

\usepackage{graphicx}

\usepackage[T1]{fontenc}

\usepackage[utf8]{inputenc}

\usepackage{microtype}

\usepackage{inconsolata}

    \makeatletter
\def\@fnsymbol#1{\ensuremath{\ifcase#1\or \dagger\or \ddagger\or
   \mathsection\or \mathparagraph\or \|\or **\or \dagger\dagger
   \or \ddagger\ddagger \else\@ctrerr\fi}}
    \makeatother

\title{Why Are Positional Encodings Nonessential for Deep Autoregressive Transformers? Revisiting a Petroglyph}

\author{Kazuki Irie\\
Department of Psychology\\ Harvard University, Cambridge MA, USA\\
  \texttt{kirie@g.harvard.edu}
}

\begin{document}
\maketitle

\begin{abstract}
Do autoregressive Transformer language models \textit{require} explicit positional encodings (PEs)?
The answer is \textit{`no'} provided they have more than one layer---they can distinguish
sequences with permuted tokens without the need for explicit PEs.
This follows from the fact that a cascade of (permutation invariant) set processors can collectively exhibit sequence-sensitive behavior in the autoregressive setting.
This property has been known since early efforts (contemporary with \mbox{GPT-2}) adopting the Transformer for language modeling \citep{irie19:trafolm}.
However, this result does not appear to have been well disseminated, leading to recent rediscoveries.
This may be partially due to a sudden growth of the language modeling community \textit{after} the advent of GPT-2/3, but perhaps also due to the lack of a clear explanation in prior work, despite being commonly understood by practitioners in the past.
Here we review the long-forgotten explanation why explicit PEs are nonessential for \textit{multi-layer} autoregressive Transformers (in contrast, \textit{one-layer} models require PEs to discern order information of their inputs), as well as the origin of this result, and hope to re-establish it as a common knowledge.
\end{abstract}

\section{Introduction}
\label{sec:intro}
The field of language modeling has seen new waves of interest after the promising results of \mbox{GPT-2} \citep{gpt2}, impressive capabilities of GPT-3 \citep{gpt3}, and unprecedented versatility of ChatGPT and GPT-4 \citep{bubeck2023sparks, achiam2023gpt}, manipulating human languages in a way no machine has ever before.

\begin{figure*}[t]
\begin{center}

\includegraphics[width=1.6\columnwidth]{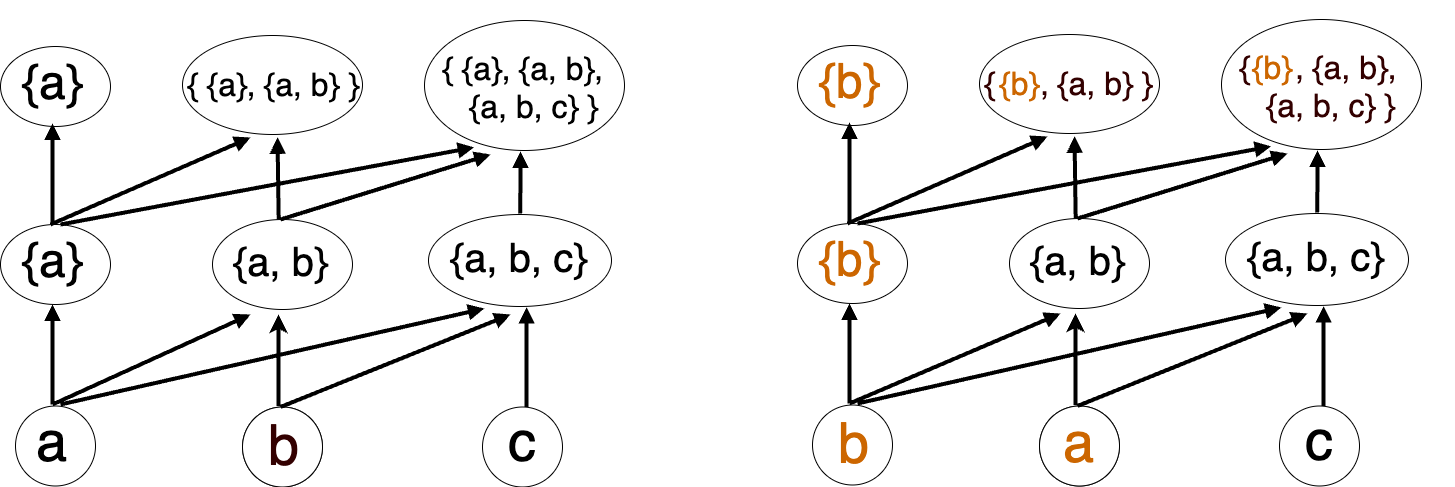}
\qquad
\caption{\textit{A cascade of set processors can behave as a sequence processor}. An illustration of autoregressive Transformers without explicit positional encodings for two input sequences: (a, b, c) and its permutation (b, a, c).
The color in the right diagram highlights the differences in terms of context seen by each layer at each position (expressed as a set).
With two (or more) layers, 
as soon as an input is different at one position (here at the first position), autoregressive Transformers see different contexts for all later positions at the top layer.
In contrast, \textit{one-layer} models can not distinguish between these two input sequences (in this example, as soon as the second token is fed) even though they are strictly speaking not permutation invariant (since the context at the first position is different). Here we only swap the two first tokens; with a more ``complex'' permutation, one-layer models may see different contexts at more positions but in all cases, they systematically fail to do so in the last step.
}
\label{fig:main}
\end{center}
\end{figure*}

About a decade before the current ``Large Language Model era'' or LLM-era\footnote{\label{foot}Here by ``LLM-era'' we roughly refer to the time after GPT-3.
The term ``petroplyph'' in the title is a hyperbole with a double meaning: first, it highlights that results from the pre-LLM era are now largely regarded as ``prehistoric'' and they are overlooked; 
second, more specifically to the positional encoding result discussed here,
figures similar to Figure \ref{fig:main} were frequently sketched in old notes and during whiteboard discussions from that time, but such a figure was not included in the 4-page Interspeech paper \citep{irie19:trafolm}.}, neural language modeling research had also seen a smaller but significant growth after Tomáš Mikolov's breakthrough results with recurrent neural network language models \citep{MikolovKBCK10,mikolov2011extensions}.
This had made neural language modeling \citep{nakamura1988study,elman1989structured,SchmidhuberH94,BengioDV00} a popular research topic, particularly among speech recognition and machine translation researchers as these two fields used to be the major application areas at the time when language models were not yet a standalone system--- they were merely a component in a larger system with a specialized application \citep{jelinek1975design, brown88}, except in certain visionary work \citep{SutskeverMH11}.

When the Transformer encoder-decoder architecture was shown to be successful for machine translation \citep{trafo}, several works investigated its application to build conventional (i.e., autoregressive) \textit{language models} using the decoder, e.g., \citet{liu2018generating,gpt1,al2018character,dai2019transformerxlacl,baevski2018adaptive}, or non-autoregressive \textit{models of language} using the encoder, e.g., \citet{devlin2019bert};
producing many methods and practical knowledge for optimizing Transformers to language modeling, concurrently to GPT-2 \citep{gpt2}.

While the recent surge of interest in language modeling has been very exciting for the field, it has also led to some discontinuities, e.g., certain common knowledge and results from pre-LLM studies appear to have been lost amid this rapid growth.

This short review focuses on one of such results, namely the property that \textit{multi-layer autoregressive Transformer language models can process sequences without explicit positional encodings} \citep{irie19:trafolm}.
In fact, it is often argued that positional encodings are necessary for Transformers, because the self-attention operation \citep{trafo,cheng16, ParikhT0U16, lin2017structured} is permutation invariant.
There is a flaw in this deduction: it overlooks the behavior of multi-layer (i.e., cascaded) self-attention in the autoregressive setting.
Here we provide a simple explanation of this result (Figure \ref{fig:main} and Sec.~\ref{sec:results}), which, although known to language modeling practitioners of the pre-LLM era, was never published (to the best of our knowledge). We also refer back to early work on this property (Sec.~\ref{sec:review}).

\section{Background: Self-Attention}
\label{sec:back}
Following the original definition \citep{trafo}, one Transformer ``layer'' consists of two sub-layers: a self-attention layer and a feedforward block.
Given that a typical feedforward block processes information at each position/time step exclusively,
the self-attention layer is the only sequence processing component of the Transformer layer.

\textbf{Autoregressive Self-Attention.} 
Let $d$ and $T$ denote positive integers.
An autoregressive self-attention layer transforms an input sequence $\{ \vx_t \}_{t=1}^T, \vx_t \in \mathbb{R}^{d}$
to an output sequence $\{ \vy_t \}_{t=1}^T, \vy_t \in \mathbb{R}^{d}$ as follows:
\begin{eqnarray}
\vq_t, \vk_t, \vv_t &=& \mW_q \vx_t, \mW_k \vx_t, \mW_v \vx_t \label{eq:proj} \\
\mK_t &=& \big[\mK_{t-1}, \vk_t] \in \mathbb{R}^{d \times t} \\
\mV_t  &=& \big[\mV_{t-1} , \vv_t ] \in \mathbb{R}^{d \times t} \\
\vy_t &=& \mV_t \softmax(\mK_t^{\top} \vq_t) \label{eq:trafo_softmax}
\end{eqnarray}
where $\mW_q, \mW_k, \mW_v \in \mathbb{R}^{d \times d}$ are trainable weight matrices, $[\mA, \va]$ denotes the concatenation of vector $\va$ to matrix $\mA$ which increments the time dimension ($\mK_0$ and $\mV_0$ are initially empty), and
$\softmax$ is along the time dimension.
We omit the $1/\sqrt{d}$ scaling inside $\softmax$, as well as the output projection, which are irrelevant for our discussion.

While the equations above accurately describe the model conceptually, self-attention is also often expressed in the following \textit{matrix form} that better reflects the possibility to parallelize computation over the time axis during training. By denoting the input as $\mX = [x_1, ..., x_T] \in \mathbb{R} ^ {d \times T}$ ($\mX_i = x_i$) and the output as $\mY = [y_1, ..., y_T] \in \mathbb{R} ^ {d \times T}$, it yields:
\begin{align}
 \mQ, \mK, \mV &= \mW_q \mX , \mW_k \mX , \mW_v \mX \label{eq:mask1} \\
\mY &= \mV \softmax(\mM \odot (\mK^{\top} \mQ)) \label{eq:mask2}
\end{align}
where $\mM \in \mathbb{R}^{T\times T}$ is the so-called \textit{attention mask}.
For these equations to be equivalent to Eqs.~\ref{eq:proj}-\ref{eq:trafo_softmax} above, i.e., for autoregressive self-attention, $\mM$ is set to be the upper triangular matrix, i.e., $\mM_{i,j} =1$ if $i \leq j$ and $\mM_{i,j} =-\infty$ otherwise.

We also denote by $\mY = \SelfAttention(\mX, \mM)$ the overall self-attention operation given input $\mX$ and mask $\mM$, grouping Eqs.~\ref{eq:mask1}-\ref{eq:mask2}.

\textbf{Non-Autoregressive Self-Attention.} 
The same equations (Eqs.~\ref{eq:mask1}-\ref{eq:mask2}) can also express \textit{non-autoregressive} self-attention by removing the mask $\mM$, i.e., by setting $\mM_{i,j}=1$ for all $i,j \in \{1, ..., T\}$. We denote such $\mM$ as $\mM=\mathbbm{1}$.

\textbf{Positional Encodings.} 
When positional encodings \citep{GehringAGYD17,trafo} are used, a vector representing discrete position $t$ is added to input token $\vx_t$.
The exact choice of PE design is irrelevant to our discussion.

\section{Main Results on Positional Encodings}
\label{sec:results}
The goal of this short review is to provide
a summary of results on the (non)necessity of positional encodings for different Transformer model variations with comprehensible explanations, and to
discuss the original references (Sec.~\ref{sec:review}).

\textbf{Definitions.} We first define two key properties:

(1) A sequence processor $f: \mathbb{R}^{d\times T} \rightarrow \mathbb{R}^{d\times T}$ is said to be \textit{permutation invariant} when for any input $\mX \in \mathbb{R}^{d\times T}$, and its arbitrary permutation along the time/token axis $\mX' \in \mathbb{R}^{d\times T}$, $f(\mX)=f(\mX')$.

(2) $f$ is \textit{fully position-sensitive} when for any inputs $\mX, \mX' \in \mathbb{R}^{d\times T}$, if $\mX_i \neq \mX'_i$, then $f(\mX)_j \neq f(\mX')_j$ for all $j \in \{i, ..., T\}$; meaning that as soon as one input is different at position $i$, $f$ produces different outputs at all the ``future'' positions $j \geq i$.

\textbf{The necessity of using explicit PEs}
is tied to the model's capability to distinguish between permutated sequences\footnote{Once it is clear that the model can distinguish between permuted sequences, 
there is no reason to introduce extra explicit PEs. A common wisdom is to let the model learn to use the positional signals on its own.
For example, it is rather unnatural to add extra PEs to recurrent neural networks (RNNs).\looseness=-1}, which can be characterized using the definitions above.
(1) If a sequence processor $f$ is \textit{permutation invariant}, positional encodings are needed.
(2) If $f$ is \textit{fully position-sensitive}, positional encodings are not needed.
(3, Remark) As we will see, strictly speaking, being \textit{permutation non-invariant} alone is not enough to conclude on the non-necessity of positional encodings (as all positions matters); therefore, we additonally introduce the property of being \textit{fully position-sensitive}.

We present the main results in the form of question/answer pairs as follows.

\begin{quest}[Back to basics]
\textit{Why are positional encodings \textbf{needed} for \textbf{non-autoregressive} Transformers?}
\end{quest}

This is because non-autoregressive self-attention is \textit{permutation invariant}, i.e., for any input $\mX$, and its arbitrary permutation along the time/token axis $\mX'$, $\SelfAttention(\mX, \mathbbm{1})=\SelfAttention(\mX', \mathbbm{1})$.

We can straightforwardly check this by directly looking at Eqs.~\ref{eq:mask1}-\ref{eq:mask2}.
Without the mask, i.e., $\mM=\mathbbm{1}$, keys and queries from all positions interact regardless of their positions to yield attention scores (i.e., $\softmax(\mK^{\top} \mQ) \in \mathbb{R}^{T \times T}$ in Eq.~\ref{eq:mask2}), which are used to compute weighted average, which is commutative, of values.

\begin{quest}[Knowledge Bias]
\textit{Why are positional encodings believed to be crucial for Transformers by ``default'' in the first place?}
\end{quest}

This is partly because the standalone self-attention is a permutation-invariant set operation (this is the case, even in the autoregressive case, if we ignore the edge case of the first token; as illustrated in Figure \ref{fig:main}).

Also, the explanation for the necessity of positional encodings in the original paper was the fact that the ``\textit{model contains no recurrence and no convolution}'' \citep{trafo}. As we'll see below, this explanation is incomplete, but if one assumed this to be true, it would imply that the autoregressive self-attention also requires positional encodings (we emphasize that this is not true).

\begin{quest}
\textit{Why are positional encodings \textbf{nonessential} for \textbf{multi-layer} \textbf{autoregressive} Transformers?}
\end{quest}

This is because multi-layer autoregressive Transformers are \textit{fully position-sensitive}.

A simple method to check this is to compare the model outputs when we feed two sequences that are permutations of each other.
While we could also provide a mathematical proof here, this can better be visualized as in Figure \ref{fig:main}: even for this extreme case where we feed two sequences that only differ from each other by a permutation of their two first tokens, the multi-layer model
sees different contexts at all positions at the top layer.
Essentially, we can build a sequence processor by cascading multiple set processors.

\begin{quest}
\textit{Does the multi-layer autoregressive Transformer language models effectively learn to use positional signals in practice?}
\end{quest}

For this question, we refer to \citet{irie19:trafolm} which demonstrated good general performance of multi-layer autoregressive Transformer language models without PEs and provided visualization of attention weights (see figures in \citet{irie19:trafolm}).
They reported that, interestingly, the first attention layer mainly attends to the new input,
while the second layer uniformly attends to the context.
Uniform attention in early layers is intuitively good as it allows the model to grasp all the available context, which is crucial to distinguish similar sequences (as illustrated in Figure \ref{fig:main}).

Finally, being non-essential does not imply that some sophisticated extra positional encodings may not improve Transformer language models, we discuss corresponding references in Sec.~\ref{sec:review}.

\begin{quest}
\textit{Why are positional encodings \textbf{needed} for \textbf{one-layer} \textbf{autoregressive} self-attention?}
\end{quest}
This is also well illustrated in Figure \ref{fig:main} by looking at the first layer.
Depending on the specific permutation, one-layer model's outputs at some positions are sensitive to the input permutation, but the output at the last position (when the entire sequence is seen) is the same for any permutations;
implying that they are not \textit{fully position-sensitive}.

\section{An Intriguing Linear Transformer Case}
\label{sec:lt}
Here we discuss an \textit{intriguing} case of 
linear Transformers \citep{katharopoulos2020transformers, schmidhuber1992learning, schlag2021linear}.
One representative example of linear Transformers can be obtained by simply removing $\softmax$ in Eq.~\ref{eq:trafo_softmax}.
The resulting model can be equivalently expressed as the following fast weight programmer (see Appendix \ref{sec:appendixa}):
\begin{align}
\tag{\ref{eq:proj}}
\vq_t, \vk_t, \vv_t &= \mW_q \vx_t, \mW_k \vx_t, \mW_v \vx_t \\
\label{eq:update}
\mW_t &= \mW_{t-1} +\vv_t \otimes \vk_t \\
\vy_t  &= \mW_t \vq_t \label{eq:get}
\end{align}
where $\otimes$ denotes outer product and $\mW_0=0$.
Because of the state update rule of Eq.~\ref{eq:update} giving an impression of ``recurrence'' (as in the title of \citet{katharopoulos2020transformers}),
it may not be immediately clear if this model requires PEs.
In reality, this is not ``true recurrence'' but a degenerated one with a recurrent transition matrix reduced to identity (for further discussions, we refer to \citet{irie2021going,irie2023practical,merrill2024illusion}).
Since this model is equivalent to the autoregressive self-attention layer discussed in Sec.~\ref{sec:results}, it inherits the same properties, i.e., PEs are needed for one-layer models, while they are nonessential for multi-layer models.

\section{Literature Review}
\label{sec:review}

To the best of our knowledge, \citet{ShenZLJPZ18} were the first to use non-symetric ``attention masks'' (Eq.~\ref{eq:mask2}) to encode positional information for neural networks whose sequence processing ability solely relies on the attention mechanism.
While the above work does not specifically discuss autoregressive language models (LMs),
their insights about masking could have been directly extended to answer the question whether positional encodings are needed for autoregressive Transformer LMs.

\citet{irie19:trafolm} empirically demonstrated that multi-layer autoregressive Transformer LMs perform well without positional encodings.
To be more specific, the corresponding ablation study was conducted for 12, 24, and 42 layer models; while other deeper models (up to 112 layers) were also trained without PEs.
This result was later rediscovered/confirmed by \citet{avivRPIL22}\footnote{Many recent papers inaccurately attribute the origin of this result (see Sec.~\ref{sec:meta} for examples and further discussions).} and \citet{KazemnejadPRDR23}.
\citet{irie19:trafolm} argued that the autoregressive setting itself encodes the positional information due to increasing context over time, which directly connects to \citet{ShenZLJPZ18}'s argument (while \citet{irie19:trafolm} failed to cite it).
Also, while they specifically state that the results are valid for the \textit{multi-layer} models, they did not explicitly discuss the \textit{one-layer} case (see Footnote \ref{foot} for further comments).
This no-positional encoding scheme has been immediately adopted in other works on speech recognition; e.g., \citet{zeyer2019:trafo-vs-lstm-asr} removed PEs from the decoder of their encoder-decoder speech recognizer.
A popular open-source speech toolkit, ESPnet \citep{watanabe2018espnet} also integrated Transformer LMs without PEs as part of their standard recipe.

\citet{LeeLKKCT19} discussed permutation invariance of non-autoregressive self-attention;
\citet{tsai2019} extended this discussion and showed that autoregressive self-attention is not permutation invariant. However, as discussed above, permutation invariance alone is not enough to conclude on the necessity of PEs (as shown in Sec.~\ref{sec:results}, one-layer autoregressive models are not \textit{permutation invariant} but also not \textit{fully position-sensitive} and require PEs).\looseness=-1

\textbf{Length Generalization.}
Empirically, whether removal of PEs yields performance improvements depends on the specific setting.
\citet{irie19:trafolm} reported general performance gain by removing absolute/sinusoidal PEs for Transformer LMs trained on books, for various numbers of layers.
In constrast, \citet{avivRPIL22} and \citet{ScaoWHBBBEMPPRS22} reported slight degradation.

Nevertheless, one of the common benefits of removing PEs is the improved length generalization.
\citet{BhattamishraAG20} showed that Transformers without PEs can generalize on certain formal languages with test sequences that are longer than the training ones.
\citet{KazemnejadPRDR23}
showed that LMs without PEs yields the best length generalization performance on reasoning-related tasks compared to sophisticated positional encoding methods.
\citet{schlag2021linear} successfully trained deep linear Transformers without PEs (Sec.~\ref{sec:lt}) by carrying context across training batches to enable them to process arbitrarily long sequences.\looseness=-1

Regarding length generalization of Transformers with \textit{non-}autoregressive self-attention, we refer to, e.g.,  \citet{csordas2021devil,csordas2021neural}.\looseness=-1

Finally, the main scope of our discussion is the (non)necessity of PEs at the conceptual level. In practice, it is often useful to augment Transformers with a certain type of PEs, especially, relative PEs \citep{su2024roformer, ShawUV18, dai2019transformerxlacl} to improve their performance; 
designing practically useful positional encodings for Transformers remains an ongoing research. The perspective developed in this work also opens up the question whether explicit PEs could also enhance other sequence processors, e.g., modern linear RNNs \citep{Bradbury17,LeiZWDA18,indRNN, BalduzziG16,gu2023mamba,QinYZ23} even when they are inherently capable of encoding positions.

\section{Conclusion}
\label{sec:ccl}
We provide a didactic explanation of why positional encodings are nonessential for multi-layer autoregressive Transformers---an explanation that was well-known among the pre-LLM language modeling practitioners but has not been formally published.
We also review the literature related to this result in the hope of correcting potential misconceptions and enhancing our collective knowledge.
\looseness=-1

\section{Excursion: Metascience Perspectives}
\label{sec:meta}
Beyond the technical scope of this work, it is also interesting, through the lens of metascience, to observe how easily the misconception about the origin of the ``no-positional encoding'' result has propagated in the current machine learning community (c.f.~the references in Sec.~\ref{sec:review}).
Many recent papers refer to this no-PE result as ``recent'' findings.
Here are some example quotes for illustration:

\begin{itemize}

\item Flamingo: a Visual Language Model for Few-Shot Learning \citep{alayrac2022flamingo}: \textit{``recent work has shown that such disambiguation is still possible implicitly through the causal attention mechanism [36] (Haviv et al. 2022).''}

\item Transformers Learn Shortcuts to Automata \citep{liu2022transformers}: \textit{``Note that removing positional encoding does not mean having no position information, since the use of the causal mask implicitly encodes the position, which is also noted in Bhattamishra et al. (2020) and concurrent work by Haviv et al. (2022).''}
 \vspace{-1mm}

\item Challenges and Applications of Large Language Models \citep{kaddour2023challenges}: \textit{``Surprisingly, Haviv et al. [192] find that causal LLMs without positional encodings are competitive compared to models with positional encodings and accredit this success to the causal attention mask leaking positional information into the model.''}
\vspace{-1mm}

\item Code Llama: Open Foundation Models for Code \citep{roziere2023code}: \textit{``Recent work suggests that causal models do not require an explicit encoding of position information (Haviv et al., 2022; Kazemnejad et al., 2023)''}
\vspace{-3mm}

\item A Phase Transition between Positional and Semantic Learning in a Solvable Model of Dot-Product Attention \citep{cui2024phase}: \textit{``While some transformers can leverage implicit positional information through causal masks in training (Haviv et al., 2022; Sinha et al., 2022; Kazemnejad et al., 2023)''}

\end{itemize}

We speculate that this is partially due to the influence of social media advertising the version-1 preprint. More concretely, neither of the recent papers on no-PEs, \citet{avivRPIL22} and \citet{KazemnejadPRDR23}, referred to the pre-LLM work that discussed the same result (Sec.~\ref{sec:review}) in their first version (arXiv v1).
Once such versions are widely advertised (to a relatively new LLM community), it seems too late to include the earlier work in a later version after a few months, unless the discovered priority issue and the correction for potential misinformation are equally advertised. 

A contrastive/positive example is the case of ``recurrent dropout'': early versions of \citet{zaremba2014recurrent}  did not cite prior work by \citet{pham2014dropout}; but after discovering \citet{pham2014dropout}, \citet{zaremba2014recurrent} updated the paper with the following statement in Page 1:
\textit{``Independently of our work, Pham et al. (2013) developed the very same RNN regularization method and applied it to handwriting recognition. We rediscovered this method and demonstrated strong empirical results over a wide range of problems.''} (and there used to be a note on their website encouraging people to cite \citet{pham2014dropout} instead of theirs).

These data points may contribute to future work in metascience and cognitive psychology of scientific referencing.

\section*{Limitations}
While our literature review reflects the authors' best efforts and knowledge, we acknowledge the possibility that prior work addressing the nonessentiality of positional encodings in autoregressive Transformers may exist.\looseness=-1

Also, here we only focused on the specific topic of positional encodings for autoregressive Transformer language models.
There are other similar cases,
including the discussion on methods to manage/reduce the size of key-value memory storage (the so-called ``KV-cache'') in autoregressive Transformers (c.f., \citet{irie2020much} and \citet{LiuDLWXXKS23, ge2023model}); or the strategy to build mixture-of-experts language models by pre-training component/expert language models independently in parallel (c.f., \citet{irie18:radmm} and \citet{li2022branch, sukhbaatar2024branch}).
Further discussion is beyond the scope of this work.

\bibliography{main}

\appendix

\section{Reminder: Derivation of the dual form of linear Transformers}
\label{sec:appendixa}

Here we briefly review the derivation \citep{katharopoulos2020transformers, schlag2021linear, ba2016using} connecting the fast weight programmer of Sec.~\ref{sec:lt} and its attention form (Eqs.~\ref{eq:proj}-\ref{eq:trafo_softmax}).
Starting from Eq.~\ref{eq:get}, and by using the definition of $\mW_t$ from Eq.~\ref{eq:update}, we obtain:
\begin{align}
\vy_t  &= \mW_t \vq_t \\
& = \left(\sum_{\tau=1}^t \vv_{\tau} \otimes \vk_{\tau}\right) \vq_t \\
& = \sum_{\tau=1}^t \vv_{\tau} \vk_{\tau}^{\top} \vq_t \\
& = \mV_t \mK_t^{\top} \vq_t 
\end{align}
where the definitions of $\mK_t$ and $\mV_t$ are as in Sec.~\ref{sec:back}.

The last equation is effectively Eq.~\ref{eq:trafo_softmax} without $\softmax$.

Note that this relation is analogous to the famous \textit{duality} that connects the perceptron to kernel machines \citep{aizerman1964theoretical,irie2022dual}.

\end{document}

%% file: math_commands.tex
\usepackage{amsmath,amsfonts,bm}

{}
{}
{}
{}

\def\eqref#1{equation~\ref{#1}}

\def\1{\bm{1}}

\def\va{{\bm{a}}}

\def\vk{{\bm{k}}}

\def\vq{{\bm{q}}}

\def\vv{{\bm{v}}}

\def\vx{{\bm{x}}}
\def\vy{{\bm{y}}}

\def\mA{{\bm{A}}}

\def\mK{{\bm{K}}}

\def\mM{{\bm{M}}}

\def\mQ{{\bm{Q}}}

\def\mV{{\bm{V}}}
\def\mW{{\bm{W}}}
\def\mX{{\bm{X}}}
\def\mY{{\bm{Y}}}

\DeclareMathAlphabet{\mathsfit}{\encodingdefault}{\sfdefault}{m}{sl}
\SetMathAlphabet{\mathsfit}{bold}{\encodingdefault}{\sfdefault}{bx}{n}

\newcommand{\softmax}{\mathrm{softmax}}
\newcommand{\SelfAttention}{\mathrm{SelfAttn}}

